\pdfoutput=1
\documentclass[runningheads]{llncs}
\usepackage{graphicx}
\usepackage[hidelinks]{hyperref}
\usepackage{color}
\usepackage{microtype}
\usepackage{mathtools}

\newcommand{\limp}{\mathbin{\Rightarrow}}
\newcommand{\liff}{\mathbin{\Leftrightarrow}}
\newcommand{\lnand}{\mathbin{\uparrow}}
\newcommand{\lnor}{\mathbin{\downarrow}}
\newcommand{\lxor}{\mathbin{\underline{\lor}}}

\newcommand{\att}{\textit{a}}
\newcommand{\true}{\text{true}}
\newcommand{\false}{\text{false}}

\DeclareMathOperator{\doesNotChange}{doesNotChange}
\DeclareMathOperator{\adj}{adj}
\DeclareMathOperator{\notEqualOrAdjacentToBlank}{notEqualOrAdjacentToBlank}
\DeclareMathOperator{\oneTileMoved}{oneTileMoved}
\DeclareMathOperator{\boardSolved}{boardSolved}

\begin{document}

\title{Effective problem solving using SAT solvers}

\author{Curtis Bright\inst{1,2} \and
J\"urgen Gerhard\inst{2} \and
Ilias Kotsireas\inst{3} \and
Vijay Ganesh\inst{1}}

\authorrunning{C.~Bright, J.~Gerhard, I.~Kotsireas, V.~Ganesh}

\institute{University of Waterloo, Waterloo, Canada \and
Maplesoft, Waterloo, Canada \and
Wilfrid Laurier University, Waterloo, Canada}

\maketitle

\begin{abstract}
In this article we demonstrate how to solve a variety of problems and puzzles
using the built-in SAT solver of the computer algebra system Maple.
Once the problems have been encoded into Boolean logic,
solutions can be found (or shown to not exist)
automatically, without the need to implement any search algorithm.
In particular, we describe how to solve the $n$-queens problem, 
how to generate and solve Sudoku puzzles, how to solve logic puzzles
like the Einstein riddle,
how to solve the 15-puzzle, how to solve the maximum clique problem,
and finding Graeco-Latin squares.
\end{abstract}

\section{Introduction}\label{sec:introduction}

\begin{quote}
``\dots it is a constant source of annoyance when you come up with a clever special algorithm which then gets beaten by translation to SAT.''
\\\mbox{}\hfill---Chris Jefferson
\end{quote}

The satisfiability (SAT) problem is to determine if a given Boolean expression can be
satisfied---is there some way of assigning true and false to its variables that makes the whole
formula true?  Despite at first seeming disconnected
from most of the kinds of problems that mathematicians care about
we argue in this paper that it is in the interests of mathematicians to
have a familiarity with SAT solving and encoding problems in SAT.
An immense amount of effort 
over the past several decades has produced SAT solvers that are not only practical
for many problems but are actually the fastest known way of
solving an impressive variety of problems such as software and hardware verification
problems~\cite{biere2009handbook}.  They have also recently been used to resolve
long-standing mathematical conjectures~\cite{heule2017solving}
and construct large combinatorial designs~\cite{bright2018sat}.

Since 2018, the computer algebra system Maple has included
the award-winning SAT solver MapleSAT~\cite{liang2017empirical} as its built-in SAT solver.
This solver can be used through the \texttt{Satisfy} command
of the \texttt{Logic} package.  \texttt{Satisfy} returns a satisfying assignment
of a given Boolean expression (if one exists)
or \texttt{NULL} if no satisfying assignment exists.
In this paper we demonstrate 
through a number of detailed examples
how \texttt{Satisfy} can be an effective and efficient
way of solving a variety of problems and puzzles.

Very little prerequisites are necessary to understand this paper;
the main necessary background is a familiarity with Boolean logic
which we outline in Section~\ref{sec:background}.
We then present effective solutions to the $n$-queens problem (Section~\ref{sec:nqueens}),
logic puzzles like the Einstein riddle (Section~\ref{sec:logic}),
Sudoku puzzles (Section~\ref{sec:sudoku}),
Euler's Graeco-Latin square problem (Section~\ref{sec:latin}),
the maximum clique problem (Section~\ref{sec:clique}),
and the 15-puzzle (Section~\ref{sec:15puzzle}).
In each case we require no knowledge of any of the special-purpose
search algorithms that have been proposed to solve these problems;
once the problems have been encoded into Boolean logic they are automatically
solved using Maple's \texttt{Satisfy}.

All of the examples discussed in this paper were implemented and run in Maple 2018 and
Maple 2019.  Due to space constraints we have not included our code
in this paper, but Maple worksheets containing complete implementations
have been made available online through the Maple Application Center~\cite{brightapps}.

\section{Background}\label{sec:background}

A basic understanding of Boolean logic is the only prerequisite necessary to understand
the solutions described in this paper.
One of the main advantages of Boolean logic (but also one of its main disadvantages)
is its simplicity: each variable can assume only
one of two values denoted by $\true$ and $\false$.
Boolean expressions consist of variables joined by Boolean operators.  The most common
Boolean operators (and the ones available in the \texttt{Logic} package of Maple)
are summarized in Table~\ref{tbl:operators}.

\begin{table}[b]
\begin{center}
\begin{tabular}{c@{\quad}c@{\qquad}c@{\qquad}c}
\bf Name & \bf Symbol & \bf Arity & \bf Maple Syntax \\
Negation & $\lnot$ & 1 & \texttt{\&not} \\
Conjunction & $\land$ & $n$-ary & \texttt{\&and} \\
Disjunction & $\lor$ & $n$-ary & \texttt{\&or} \\
Implication & $\limp$ & 2 & \texttt{\&implies} \\
Biconditional & $\liff$ & 2 & \texttt{\&iff} \\
Alternative denial & $\lnand$ & $n$-ary & \texttt{\&nand} \\
Joint denial & $\lnor$ & $n$-ary & \texttt{\&nor} \\
Exclusive disjunction & $\lxor$ & $n$-ary & \texttt{\&xor} \\
\end{tabular}
\end{center}
\caption{The Boolean logical operators available in Maple.}\label{tbl:operators}
\end{table}

The $\lor$ (or), $\land$ (and), and $\lnot$ (not) operators have meanings based on their
everyday English meanings:
$x_1\lor\dotsb\lor x_n$
is true exactly when at least one $x_i$ is true,
$x_1\land\dotsb\land x_n$ is true
exactly when all $x_i$ are true,
and $\lnot x$ is true exactly when~$x$ is false.
More generally,
$x\liff y$ is true exactly when $x$ and $y$ have the same truth values,
$x\limp y$ is false exactly when $y$ is true and $x$ is false,
$x_1\lxor\dotsb\lxor x_n$ is true exactly when an odd number of $x_i$
are true, $x_1\lnand\dotsb\lnand x_n$ is true exactly when
at least one~$x_i$ is false, and $x_1\lnor\dotsb\lnor x_n$ is true exactly
when all~$x_i$ are false.

A \emph{literal} is an expression of the form $x$ or $\lnot x$ where
$x$ is a Boolean variable.  A \emph{clause} is an expression of the form
$l_1\lor\dotsb\lor l_n$ where all $l_i$ are literals.  A~\emph{conjunctive normal form} (CNF)
expression is of the form $c_1\land\dotsb\land c_n$ where all~$c_i$ are clauses.
A standard theorem of Boolean logic is that any expression can be converted
into an equivalent expression in conjunctive normal form where two expressions
are said to be \emph{equivalent} if they assume the same truth values under all
variable assignments.

The current algorithms used in state-of-the-art SAT solvers
require that the input formula be given in conjunctive normal form.
While this is convenient for the purposes of designing efficient solvers it is
not convenient for the mathematician who wants to express their problem in
Boolean logic---not all expressions are \emph{naturally} expressed in conjunctive normal form.
An advantage that Maple has over most current state-of-the-art
SAT solvers is that Maple does not require the input to be given in conjunctive normal form. 
Since the algorithms used by MapleSAT require CNF
to work properly, Maple internally converts expressions into CNF
automatically.  This is done by using a number of equivalence transformations, e.g., the expression
$x\limp y$ is rewritten as the clause $\lnot x\lor y$.

Care has been taken to make the necessary conversion to CNF efficient.
This is important because conversions that use the most straightforward
equivalence rules generally require exponential time to complete.  For example, the
Maple command \texttt{Normalize} from the \texttt{Logic} package can be used to
convert an expression into CNF.  But watch out---many expressions explode in
size following this conversion.  For example, the expression
$x_1\lxor\dotsb\lxor x_n$ when converted into CNF contains $2^{n-1}$ clauses.
The main trick used to make the conversion into CNF efficient is the \emph{Tseitin
transformation}~\cite{tseitin1970on}.  This transformation avoids the exponential blowup of the straightforward
transformations by using additional variables to derive a new formula that is satisfiable
if and only if the original formula is satisfiable.  For example, the expression
$x_1\lxor\dotsb\lxor x_n$ is rewritten as $(t\lxor x_3\lxor\dotsb\lxor x_n)\land C$ where~$t$
is a new variable and $C$ is a CNF encoding of the formula $t\liff(x_1\lxor x_2)$, namely,
\[ (\lnot x_1\lor x_2\lor t)\land(x_1\lor\lnot x_2\lor t)\land(x_1\lor x_2\lor\lnot t)\land(\lnot x_1\lor\lnot x_2\lor\lnot t) . \]
The transformation is then recursively applied to $t\lxor x_3\lxor\dotsb\lxor x_n$
(the part of the formula not in CNF) until the entire formula is in CNF.  The Maple
command \texttt{Tseitin} of the \texttt{Logic} package can be applied to
convert an arbitrary formula into CNF using this translation.
Thus, Maple offers us the convenience of not requiring encodings to be in CNF
while avoiding the inefficiencies associated with a totally unrestricted encoding.

\section{The $n$-queens problem}\label{sec:nqueens}

\begin{figure}
\begin{minipage}{0.7\linewidth}
\centering\includegraphics[scale=0.25]{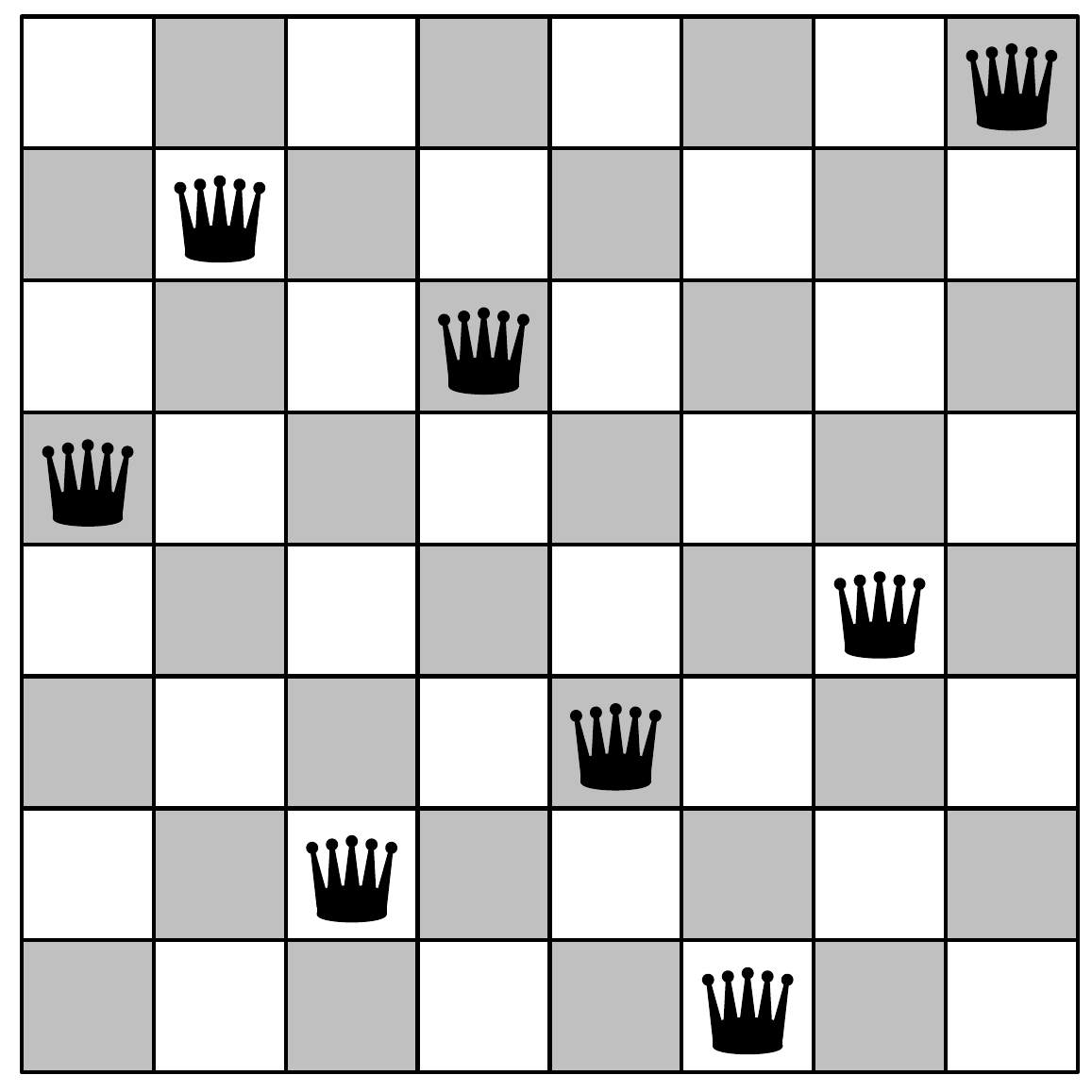}
\end{minipage}
\begin{minipage}{0.1\linewidth}
\mbox{}\\[-2\baselineskip]
\begin{align*}
Q_{1, 5} &= \true \\
Q_{2, 7} &= \true \\
Q_{3, 2} &= \true \\
Q_{4, 6} &= \true \\
Q_{5, 3} &= \true \\
Q_{6, 1} &= \true \\
Q_{7, 4} &= \true \\
Q_{8, 8} &= \true
\end{align*}
\end{minipage}
\caption{A visual representation of a solution
for the 8-queens problem (left)
and the variables assigned to true in this solution
using our SAT encoding.  Following traditional chess convention,
columns are indexed left to right
and rows are indexed bottom to top.}
\label{fig:queens}
\end{figure}

The $n$-queens problem is to place~$n$ chess queens on an $n\times n$ chessboard
such that no two queens are mutually attacking (i.e., in the same row, column, or diagonal).
The problem was first proposed for $n=8$ by Bezzel in 1848 and the
first solution for general~$n$ was given by Pauls in 1874~\cite{bell2009survey}.
The problem is solvable for all $n\geq4$; a solution for $n=8$ (found in 0.015 seconds
using \texttt{Satisfy}) is shown in Figure~\ref{fig:queens}.

The $n$-queens problem is a standard example of a
constraint satisfaction problem~\cite{nadel1990representation}.
The encoding that we use for this problem uses the $n^2$ Boolean variables $Q_{x,y}$
with $1\leq x,y\leq n$ to denote if there is a queen on square $(x,y)$.
There are two kinds of constraints necessary for this problem: \emph{positive}
constraints that say that there are $n$ queens on the board and \emph{negative}
constraints that say that queens do not attack each other.  A satisfying assignment
of these constraints exists exactly when the $n$-queens problem is solvable.

Since there are $n$ rows and each row must contain a queen the positive constraints
are of the form $Q_{1,j} \lor \dotsb \lor Q_{n,j}$ for $1\leq j\leq n$.
Similarly, each column must contain a queen; these constraints are of the form
$Q_{i,1}\lor\dotsb\lor Q_{i,n}$ for $1\leq i\leq n$.
The negative constraints say that if $(x,y)$ contains a queen then all squares attacked by $(x,y)$
do not contain a queen. 
These constraints are represented in Boolean logic by
$Q_{x,y}\limp\lnot(A_1\lor\dotsb\lor A_k)$
where $\{A_1,\dotsc,A_k\}$ are the variables ``attacked'' by a queen on $(x,y)$.
In general this encoding uses $\Theta(n^2)$ constraints in order~$n$.
Typically \texttt{Satisfy} is able to solve each order using slightly more
time than the previous order
and the last order it can solve in under a second is $n=32$.

\section{The Einstein riddle}\label{sec:logic}

The Einstein riddle is a logic puzzle apocryphally attributed to Albert Einstein
and is often stated with the remark that it is only solvable by 2\% of the world's population.
The true source of the puzzle is unknown, but a version of it appeared in the magazine
Life International in 1962.
In the puzzle there are five houses in a row with each house a different colour and each
house owned by a man of a different nationality.  Additionally, each of the owners have
a different pet, prefer a different kind of drink, and smoke a different brand of cigarette.
Furthermore, the following information is given:
\begin{enumerate}
\item The Brit lives in the red house.
\item The Swede keeps dogs as pets.
\item The Dane drinks tea.
\item The green house is next to the white house, on the left.
\item The owner of the green house drinks coffee.
\item The person who smokes Pall Mall rears birds.
\item The owner of the yellow house smokes Dunhill.
\item The man living in the centre house drinks milk.
\item The Norwegian lives in the first house.
\item The man who smokes Blends lives next to the one who keeps cats.
\item The man who keeps horses lives next to the man who smokes Dunhill.
\item The man who smokes Blue Master drinks beer.
\item The German smokes Prince.
\item The Norwegian lives next to the blue house.
\item The man who smokes Blends has a neighbour who drinks water.
\end{enumerate}
The puzzle is: Who owns the fish?

To solve this riddle using Maple, we label the houses 1 to 5
and use the variables $S_{i,\att}$
where $1\leq i\leq 5$ and $\att$ is an attribute (one of the colours,
nationalities, pets, drinks, or cigarette brands).
For example, if $a$ is a colour then $a$ is in the set
$C\coloneqq\{\text{red}, \text{green}, \text{white}, \text{yellow}, \text{blue}\}$
and similarly for the other attribute types;
there are five distinct possible attributes for each type of attribute.
In total there are $5^2=25$ possible
values for~$a$ and $5^3=125$ variables~$S_{i,\att}$.

We know that each attribute is not shared among the five houses or their owners.
Since there are exactly five houses, each attribute must appear exactly once among the five houses.
The knowledge that each attribute appears at least once can be encoded as the clauses 
$\bigvee_{i=1}^5 S_{i,a}$
for each attribute $a$
and the knowledge that each attribute is not shared can be encoded as
$S_{i,a}\limp\lnot S_{j,a}$
where~$j$ is a house index not equal to $i$
and $a$ is an attribute.
Additionally, the fact that each house has some colour is encoded as
$\bigvee_{a\in C}S_{i,a}$
for each house index~$i$ and the knowledge that each house cannot have two colours
can be encoded as
$S_{i,c}\limp\lnot S_{i,d}$ where $i$ is a house index and $c$ and $d$ are two distinct colours
(and similarly for the other kinds of attributes).

The known facts can be encoded into logic fairly straightforwardly; for example,
the first fact can be encoded into logic as $S_{i,\text{Brit}}\limp S_{i,\text{red}}$ for house indices~$i$
and the last fact can be encoded as $S_{i,\text{Blends}}\limp(S_{i-1,\text{water}}\lor S_{i+1,\text{water}})$
for $1<i<5$ and $(S_{1,\text{Blends}}\limp S_{2,\text{water}})\land(S_{5,\text{Blends}}\limp S_{4,\text{water}})$.
Using \texttt{Satisfy} on these constraints produces the unique satisfying solution
(that includes the equations $S_{4,\text{German}}=\true$ and $S_{4,\text{fish}}=\true$,
thereby solving the puzzle) in under 0.01 seconds.

\section{Sudoku puzzles}\label{sec:sudoku}

\begin{figure}
\begin{minipage}{0.5\linewidth}
\centering\includegraphics[scale=0.25]{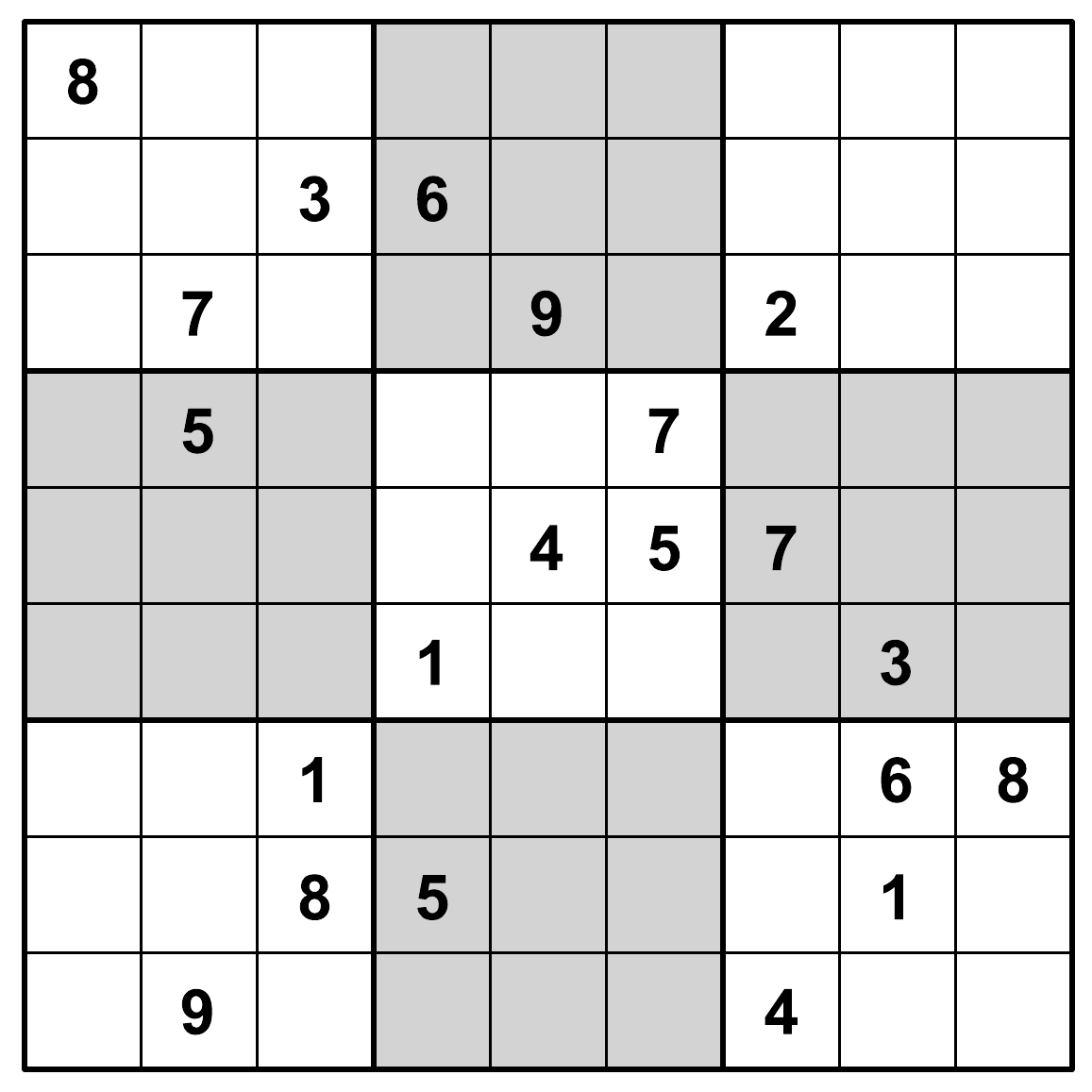}
\end{minipage}
\begin{minipage}{0.5\linewidth}
\mbox{}\\[-2\baselineskip]
\begin{align*}
S_{1,1,8} && S_{2,3,3} && S_{2,4,6} \\
S_{3,2,7} && S_{3,5,9} && S_{3,7,2} \\
S_{4,2,5} && S_{4,6,7} && S_{5,5,4} \\
S_{5,6,5} && S_{5,7,7} && S_{6,4,1} \\
S_{6,8,3} && S_{7,3,1} && S_{7,8,6} \\
S_{7,9,8} && S_{8,3,8} && S_{8,4,5} \\
S_{8,8,1} && S_{9,2,9} && S_{9,7,4}
\end{align*}
\end{minipage}
\caption{A Sudoku puzzle estimated to have a difficulty rating of 11 stars,
whereas the most challenging Sudoku puzzles that are usually published
are given 5 stars.  On the right are the starting constraints (as unit clauses)
of our encoding for this puzzle.}
\label{fig:sudoku}
\end{figure}

Sudoku is a popular puzzle that appears in many puzzle books and newspapers.
Given a 9 by 9 grid whose squares are either blank or contain a number between~1 and 9,
the objective is to fill in the blank squares in such a way that each row and column
contains exactly one digit between 1 and 9.  Additionally, each of the nine 3 by 3
subgrids which compose the grid (called blocks)
must also contain exactly one digit between 1 and 9.
Figure~\ref{fig:sudoku} contains a Sudoku puzzle designed by mathematician
Arto Inkala and claimed to be the world's hardest Sudoku~\cite{collins2012worlds}.

It is known that Sudoku can be modelled as a SAT problem~\cite{lynce2006sudoku} or a constraint satisfaction
problem~\cite{russell2016artificial}.  A straightforward encoding uses $9^3=729$ variables
$S_{i,j,k}$ with $1\leq i,j,k\leq 9$ where $S_{i,j,k}$ is true exactly when
the square $(i,j)$ contains the digit~$k$.
The rules of Sudoku state that each square must be filled with a digit between~1 and~9
and that the same digit cannot appear twice in the same row, column, or block.
The first constraint has the form $S_{i,j,1}\lor\dotsb\lor S_{i,j,9}$
for all $1\leq i,j\leq 9$
and the second constraint has the form 
$S_{i,j,k}\limp\lnot S_{i',j',k}$
for all $1\leq i,j,i',j',k\leq 9$ where
$(i,j)$ does not equal $(i',j')$
but is in the same row, column, or block as $(i,j)$.

One can also include the constraints
$S_{i,j,k}\limp\lnot S_{i,j,k'}$ for all $1\leq i,j,k,k'\leq 9$
with $k\neq k'$ that say that each square can contain
at most one digit.  However, these constraints are unnecessary
since they are logically implied by the first two constraints.  In our
tests, including these additional constraints
slightly decreased the performance of \texttt{Satisfy}.
Without the additional constraints the ``world's hardest Sudoku''
was solved in 0.25 seconds and with them it was solved in 0.33 seconds.

Additionally, we developed a method of generating Sudoku puzzles
with a unique solution using \texttt{Satisfy}.
This allowed us to write an interactive Sudoku game
where random puzzles can automatically be generated on command.
To begin, \texttt{Satisfy} is used to find a solution to the above
Sudoku constraints with an empty grid (no starting clues)
and the produced satisfying solution generates a completed Sudoku grid~$R$.
A random seed is passed to the SAT solver so that a different
solution~$R$ is generated each time; this is done by passing the following
\texttt{solveroptions} in the call to \texttt{Satisfy}:
\[ \verb|[rnd_init_act=true, random_seed=floor(1000*time[real]())]| \]
Let $R_{i,j}$ denote the $(i,j)$ the entry in the solution~$R$
where $1\leq i,j\leq9$.
The $R_{i,j}$ are randomly ordered and the first 50 entries $R_{i,j}$ are
selected as the potential starting configuration of a Sudoku puzzle.
This puzzle has the solution~$R$ by construction, though other solutions
may also exist.

To verify that the generated solution is unique, we re-run \texttt{Satisfy} with
the additional 50 unit clauses corresponding to the starting configuration
along with the constraint $\bigvee_{R_{i,j}=k}\lnot S_{i,j,k}$ which blocks the solution 
$R$.  If \texttt{Satisfy} returns another solution then we start over and find a new 
$R$ to try.  Otherwise the starting configuration forms a legal Sudoku puzzle.

Additionally, it may be the case that we can use fewer than 50 entries and still obtain
a Sudoku puzzle with a unique solution. To estimate how many entries need to be assigned
using only a few extra calls to the SAT solver we use a variant of binary search, letting
$l\coloneqq20$ and $h\coloneqq50$ be lower and upper bounds on how many
entries we will define in the puzzle.  Next, we let
$m\coloneqq\operatorname{round}((l+h)/2)$ and repeat the first step except using only the first~$m$
entries $R_{i,j}$.  If the resulting SAT instance is satisfiable then we need to use strictly more than 
$m$ entries to ensure that a unique solution exists and if the resulting SAT instance is unsatisfiable
then we can perhaps use strictly fewer than~$m$ entries.
Either way, we improve the bounds on how many entries to assign (in the former case we can update 
$l$ to~$m$ and in the latter case we can update 
$h$ to~$m$) and this step can be repeated a few times to find more precise bounds on how
many entries need to be assigned to ensure a unique solution exists.

\section{Euler's Graeco-Latin square problem}\label{sec:latin}

\begin{figure}
\begin{minipage}{0.5\linewidth}
\centering\includegraphics[scale=0.25]{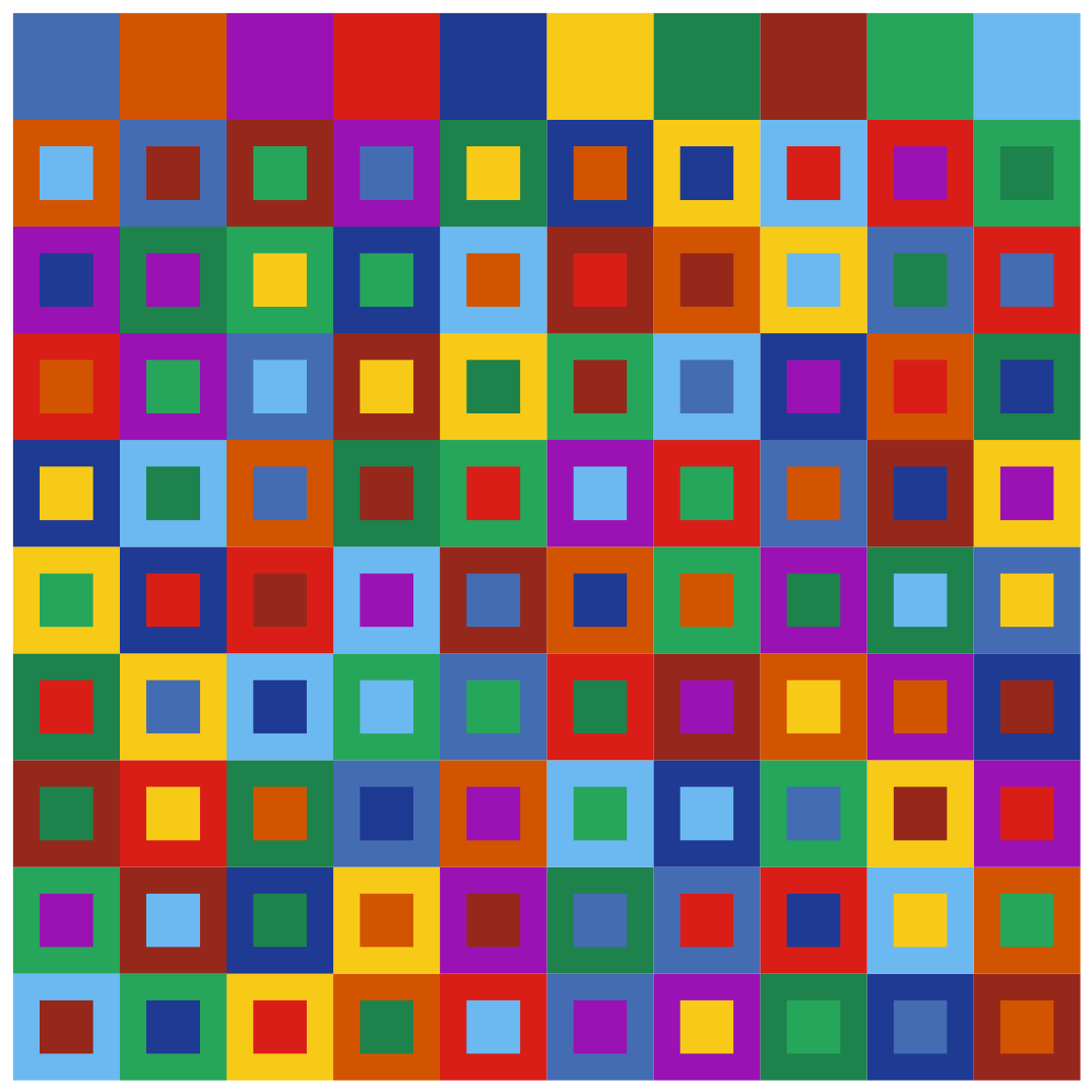}
\end{minipage}
\begin{minipage}{0.5\linewidth}
\mbox{}\\[-0.75\baselineskip]
\begin{align*}
A_{1,1,1} &= \true & B_{1,1,1} &= \true \\
A_{2,1,2} &= \true & B_{2,1,10} &= \true \\
A_{3,1,3} &= \true & B_{3,1,5} &= \true \\
A_{4,1,4} &= \true & B_{4,1,2} &= \true \\
A_{5,1,5} &= \true & B_{5,1,6} &= \true \\
A_{6,1,6} &= \true & B_{6,1,9} &= \true \\
A_{7,1,7} &= \true & B_{7,1,4} &= \true \\
A_{8,1,8} &= \true & B_{8,1,7} &= \true \\
A_{9,1,9} &= \true & B_{9,1,3} &= \true \\
A_{10,1,10} &= \true & B_{10,1,8} &= \true \\
\end{align*}
\end{minipage}
\caption{On the left is a visual representation of a Graeco-Latin
square of order~$10$ with each colour denoting a separate integer.
The entries of the second Latin square (represented by the small squares in the image)
are superimposed onto the entries of the first Latin square.
On the right are the variables corresponding
to the first column of this Graeco-Latin square that are assigned
to true using our encoding.}
\label{fig:latin}
\end{figure}

A \emph{Latin square} of order $n$ is an $n\times n$ matrix containing
integer entries between~$1$ and~$n$ such that every row and every column
contains each entry exactly once.
Two Latin squares are \emph{orthogonal} if the superposition of one over the other produces all 
$n^2$ distinct pairs of integers between~$1$ and~$n$.
A pair of orthogonal Latin squares was called a \emph{Graeco-Latin square}
by the mathematician Leonhard Euler who in 1782 used Latin characters to
represent the entries of the first square and Greek characters to represent the
entries of the second square~\cite{euler1782recherches}.
Figure~\ref{fig:latin} contains a visual representation of a Graeco-Latin square.

Euler studied the orders $n$ for which Graeco-Latin squares exist and found methods
for constructing them when $n$ was odd or a multiple of 4.
Since such squares do not exist for $n = 2$ and he was unable to find a solution for 
$n = 6$ he conjectured that Graeco-Latin squares do not exist when $n\equiv2\pmod{4}$.
Euler's conjecture became famous as he was not able to resolve it in his lifetime.

The first progress on the conjecture did not come until over a hundred
years later when in 1900 Tarry showed that
Graeco-Latin squares of order~$6$ do not exist~\cite{tarry1901probleme}.
This gave credence to Euler's conjecture
and many mathematicians thought the conjecture was true---in fact, three independent
proofs of the conjecture were published in the early 20th
century~\cite{macneish1922euler,peterson1902les,wernicke1910problem}.
In 1959--1960
Bose, Shrikhande, and Parker~\cite{bose1960further,bose1959falsity}
made explosive news (even appearing on the front page of the New York Times)
by showing that these proofs were invalid by giving explicit constructions for
Graeco-Latin squares in all orders except two and six.
As it turns out, a lot of time could have been saved if Euler had a copy of Maple---we now
show that Euler's conjecture can be automatically disproven in Maple.
With \texttt{Satisfy} we are able to construct small Graeco-Latin squares
without any knowledge of search algorithms or construction methods.

Our encoding for the Graeco-Latin square problem of order~$n$
uses the $2n^3$ variables $A_{i,j,k}$ and $B_{i,j,k}$ with $1\leq i,j,k\leq n$.
The variables $A_{i,j,k}$ will be true exactly when the $(i,j)$th entry of
the Latin square $A$ is~$k$ and $B_{i,j,k}$ will be true exactly when the $(i,j)$th
entry of the Graeco square $B$ is~$k$.

There are three kinds of constraints that specify that $(A,B)$ is a Graeco-Latin square:
Those that specify that every entry of~$A$ and~$B$ is an integer between~$1$ and~$n$,
those that specify that the rows and columns of~$A$ and~$B$ contain no duplicate entries,
and those that specify that~$A$ and~$B$ are orthogonal.
Additionally, there are constraints that are not logically necessary but help cut down
the search space.
Some work has previously been done using SAT solvers to search for special kinds of Graeco-Latin
squares~\cite{zaikin2015search}.  The encoding we use is similar but takes advantage of the
fact that Maple does not require constraints to be specified in conjunctive normal form.

First, we specify that the entries of~$A$ are well-defined, i.e.,
consist of a single integer between~$1$ and~$n$.  The constraints that say that
each entry of $A$ contains at least one integer are of the form $A_{i,j,1}\lor\dotsb\lor A_{i,j,n}$
for each index pair $(i,j)$ and the constraints that say that each entry of $A$ contains at most
one integer are of the form $A_{i,j,k}\limp\lnot A_{i,j,l}$ for each index pair $(i,j)$ and integer $k\neq l$.
Similar constraints are also used to specify that the entries of~$B$ are well-defined.

Second, we specify that~$A$ is a Latin square, i.e., all columns and rows contain distinct entries.
These have the form $A_{i,j,k}\limp\lnot A_{i',j',k}$ where $1\leq k\leq n$ and
$(i,j)\neq(i',j')$ but $(i,j)$ is in the same column or row as $(i',j')$.  Similarly, we also
specify that~$B$ is a Latin square.

Third, we specify that~$A$ and $B$ are orthogonal, i.e., for every pair $(k,l)$ there exists
some pair $(i,j)$ such that $A_{i,j,k}\land B_{i,j,l}$ holds.  These constraints are of the form
$\bigvee_{i,j=1}^n(A_{i,j,k}\land B_{i,j,l})$ for each pair $(k,l)$.

Lastly, we include some ``symmetry breaking'' constraints.  These constraints are not strictly
necessary but they shrink the search space and thereby make the search more efficient.
In general, when a search space splits into symmetric subspaces it is beneficial to add
constraints that remove or ``break'' the symmetry.
Graeco-Latin squares $(A,B)$ have a number of symmetries, in particular, a row or column
permutation simultaneously applied to $A$ and $B$ produces another Graeco-Latin square.
Also, any permutation of $\{1,\dotsc,n\}$ may be applied to the entries of either 
$A$ or $B$.

The result of these symmetries is that any Graeco-Latin square can be transformed into one
where the first row and column of $A$ has entries in ascending order (by permuting rows/columns)
and the first row of $B$ has entries in ascending order (by renaming the entries of $B$).
Thus, we can assume the constraint $\bigwedge_{i=1}^n(A_{1,i,i}\land B_{1,i,i}\land A_{i,1,i})$.
Altogether this encoding uses $\Theta(n^2)$ constraints.

Using this encoding the orders up to eight can be solved in 25 total seconds
(including 14 seconds to show that no Graeco-Latin squares exist in order six),
a Graeco-Latin square of order nine can be found in about 45 minutes, and a
Graeco-Latin square of order ten can be found in about 23 hours,
thereby disproving Euler's Graeco-Latin square conjecture.

\section{The maximum clique problem}\label{sec:clique}

\begin{figure}
\begin{minipage}{0.7\linewidth}
\centering\includegraphics[scale=0.25]{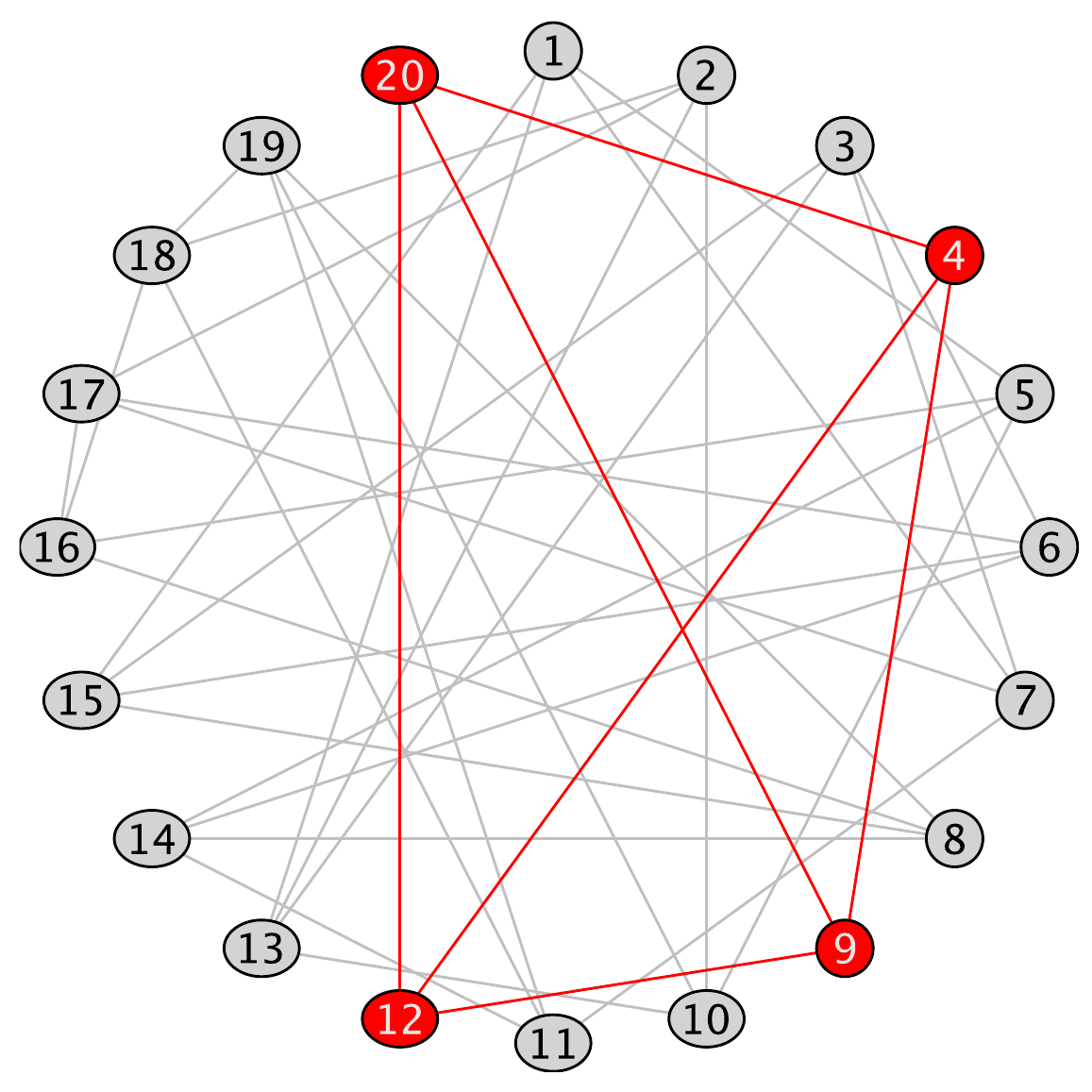}
\end{minipage}
\begin{minipage}{0.1\linewidth}
\mbox{}\\[-2\baselineskip]
\begin{align*}
x_4&=\true \\
x_9&=\true \\
x_{12}&=\true \\
x_{20}&=\true \\
s_{4,1} &= \true \\
s_{9,2} &= \true \\
s_{12,3} &= \true \\
s_{20,4} &= \true
\end{align*}
\end{minipage}
\caption{On the left is a visual representation of a graph with 20 vertices
and a highlighted clique of size 4 that was found in 0.025 seconds.
On the right are the important variables assigned to true in the assignment
returned by \texttt{Satisfy} for our encoding of the maximum clique problem for this graph.}
\label{fig:clique}
\end{figure}

The maximum clique problem is to find a clique of maximum size in a given graph.
A \emph{clique} of a graph is a subset of its vertices that are all mutually connected
(see Figure~\ref{fig:clique}).
The decision version of this problem (does a graph contain a clique of size~$k$?)
is in NP, meaning that it is easy to verify the correctness of a solution if one can be found.
By the Cook--Levin theorem~\cite{cook1971complexity} the problem
can be encoded into a SAT instance in polynomial time.
However, the reduction involves simulating
the computation of a machine that solves the maximum clique problem and is therefore
not very convenient to use in practice.  Thus, we provide a simpler encoding into Boolean logic.

Suppose the given graph~$G$ has vertices labelled $1$, $\dotsc$,~$n$ and we want to
find a clique of size~$k$ in~$G$.
Our encoding uses the variables $x_1$, $\dotsc$, $x_n$ where
$x_i$ represents that the vertex~$i$ appears in the clique we are attempting
to find.
We need to enforce a constraint that says that if $x_i$ and $x_j$
are true (for any distinct vertices $1\leq i,j\leq n$) then the edge $\{i,j\}$
exists in the graph $G$.
Equivalently, if the edge $\{i,j\}$ does not exist in the graph~$G$
then the variables $x_i$ and $x_j$ cannot both be true (for any vertices $i,j$).
In other words, for every edge $\{i,j\}$ in the complement of~$G$
we use the clause $\lnot x_i\lor\lnot x_j$.

Additionally, we need a way to enforce that the found clique is of size 
$k$.  The most naive way to encode this is
as a disjunction over all $\binom{n}{k}$ conjunctions of length $k$
on the variables $x_1$, $\dotsc$, $x_n$.  However,
this encoding is very inefficient in practice.
A cleverer encoding uses Boolean counter variables $s_{i,j}$
(where $0\leq i\leq n$ and $0\leq j\leq k$)
that represent that at least $j$ of the variables 
$x_1$, $\dotsc$, $x_i$ are assigned to true.
We know that $s_{0,j}$ will be false for $1\leq j\leq k$
and that $s_{i,0}$ will be true for $0\leq i\leq n$.
Additionally, we know that $s_{i,j}$ is true exactly when
$s_{i-1,j}$ is true or $x_i$ is true and $s_{i-1,j-1}$ is true.
This is represented by the formulas 
\[ s_{i,j} \liff (s_{i-1,j}\lor(x_i\land s_{i-1,j-1})) \qquad \text{for $1\leq i\leq n$ and $1\leq j\leq k$} \]
or in conjunctive normal form by the clauses
$\lnot s_{i-1,j}\lor s_{i,j}$, $\lnot x_i\lor \lnot s_{i-1,j-1}\lor s_{i,j}$,
$\lnot s_{i,j}\lor s_{i-1,j}\lor x_i$, and $\lnot s_{i,j}\lor s_{i-1,j}\lor s_{i-1,j-1}$.
To enforce that the found clique contains at least $k$ vertices we also assign $s_{n,k}$ to true.

To solve the maximum clique problem for a given graph $G$ we initialize~$k$ to~$3$
(assuming the graph has at least one edge, otherwise the problem is trivial)
and use the above encoding to search for a clique of size~$k$.  If such a clique
exists we increase~$k$ by~$1$ and repeat the search in this manner until a
clique of size~$k$ does not exist.  The last satisfying assignment found then provides
a maximum clique of~$G$ using an encoding with $\Theta(nk)$ variables and $\Theta(n^2)$ clauses.

\begin{table}
\begin{center}
\begin{tabular}{c@{\quad}c@{\quad}c@{\quad}c@{\quad}c@{\quad}c}
Benchmark & SAT time (sec) & Maple 2018 (sec) & Vertices & Edges & Clique size \\
\verb|brock200_2| & 23.52 & \bf 22.19 & 200 & 9876 & 12 \\
\verb|c-fat200-1| & 0.71 & \bf 0.03 & 200 & 1534 & 12 \\ 
\verb|c-fat200-2| & 2.32 & \bf 0.13 & 200 & 3235 & 24 \\ 
\verb|c-fat200-5| & \bf 9.07 & 20.63 & 200 & 8473 & 58 \\
\verb|c-fat500-1| & 4.78 & \bf 0.13 & 500 & 4459 & 14 \\ 
\verb|c-fat500-2| & 11.12 & \bf 0.82 & 500 & 9139 & 26 \\ 
\verb|c-fat500-5| & \bf 42.46 & 132.84 & 500 & 23191 & 64 \\ 
\verb|c-fat500-10| & \bf 134.42 & Timeout & 500 & 46627 & 126 \\ 
\verb|hamming6-2| & \bf 0.64 & 51.10 & 64 & 1824 & 32 \\
\verb|hamming6-4| & 0.04 & \bf 0.02 & 64 & 704 & 4 \\ 
\verb|hamming8-2| & \bf 58.51 & Timeout & 256 & 31616 & 128 \\ 
\verb|hamming8-4| & \bf 7.96 & 3393.26 & 256 & 20864 & 16 \\ 
\verb|johnson8-2-4| & \bf 0.01 & \bf 0.01 & 28 & 210 & 4 \\ 
\verb|johnson8-4-4| & \bf 0.23 & 7.80 & 70 & 1855 & 14 \\ 
\verb|johnson16-2-4| & \bf 5.62 & 642.24 & 120 & 5460 & 8 \\ 
\verb|keller4| & \bf 7.76 & 414.80 & 171 & 9435 & 11 \\ 
\verb|MANN_a9| & \bf 0.13 & 226.18 & 45 & 918 & 16 \\ 
\verb|p_hat300-1| & 11.81 & \bf 3.30 & 300 & 10933 & 8 \\ 
\verb|p_hat500-1| & 308.87 & \bf 34.60 & 500 & 31569 & 9 \\ 
\verb|p_hat700-1| & 1281.62 & \bf 169.68 & 700 & 60999 & 11 
\end{tabular}
\end{center}
\caption{A comparison of the SAT method and the \texttt{MaximumClique} function of Maple 2018
on a collection of maximum clique benchmarks with a timeout of an hour.}\label{tbl:timings}
\end{table}

This implementation was tested on the maximum clique problems from the second DIMACS
implementation challenge~\cite{johnson1996cliques}. Additionally, it was compared with Maple's
\texttt{MaximumClique} function from the \texttt{GraphTheory} package
that uses a branch-and-bound backtracking algorithm~\cite{kreher1998combinatorial}.
Of the 80 benchmarks, the SAT method solved 18 in under 3 minutes
and the branch-and-bound method solved 15.  The SAT approach was faster
in over half of the solved benchmarks and 
in one case solved a benchmark in 8 seconds that \texttt{MaximumClique}
required 57 minutes to solve (see Table~\ref{tbl:timings}).

The SAT method has been made available in Maple 2019 by using the \texttt{method=sat}
option of \texttt{MaximumClique}.  
By default the \texttt{MaximumClique} function in Maple 2019 will run the
previous method used by Maple and the SAT method in parallel and return the answer of whichever
method finishes first.  This hybrid approach is the method of choice
especially when more than a single core is available.

\section{The 15-puzzle}\label{sec:15puzzle}

\begin{figure}
\begin{minipage}{0.5\linewidth}
\centering\includegraphics[scale=0.25]{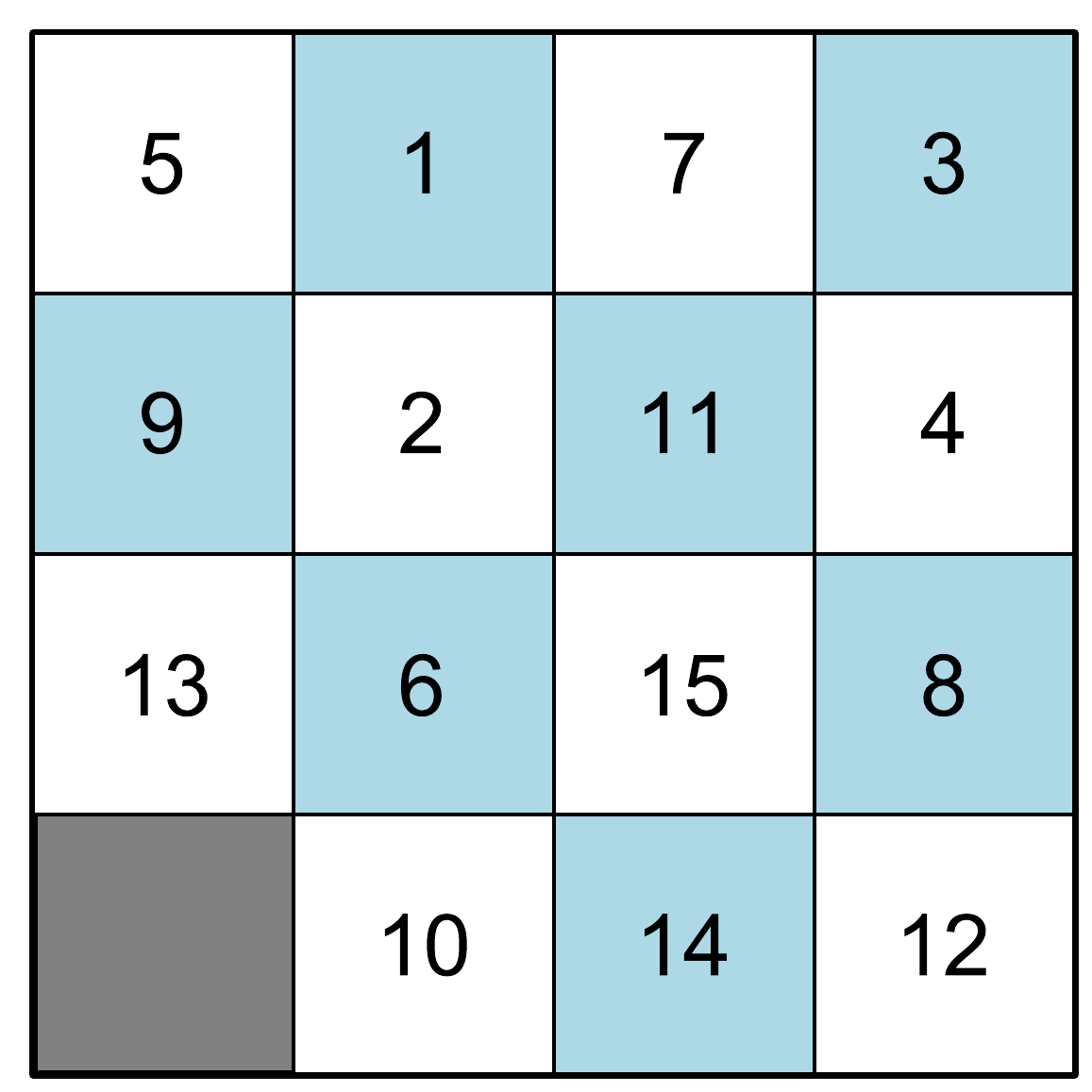}
\end{minipage}
\begin{minipage}{0.45\linewidth}
\mbox{}\\[-\baselineskip]
\[\def\arraystretch{1.2}\begin{array}{c@{\quad}c@{\quad}c@{\quad}c}
S_{1, 1, 5, 0} & S_{1, 2, 1, 0} & S_{1, 3, 7, 0} & S_{1, 4, 3, 0} \\
S_{2, 1, 9, 0} & S_{2, 2, 2, 0} & S_{2, 3, 11, 0} & S_{2, 4, 4, 0} \\
S_{3, 1, 13, 0} & S_{3, 2, 6, 0} & S_{3, 3, 15, 0} & S_{3, 4, 8, 0} \\
S_{4, 1, 16, 0} & S_{4, 2, 10, 0} & S_{4, 3, 14, 0} & S_{4, 4, 12, 0}
\end{array}\]
\end{minipage}
\caption{On the left is a visual representation of one starting configuration
of the 15-puzzle and on the right are the starting constraints (as unit clauses) for this
starting configuration in our encoding.}
\label{fig:15puzzle}
\end{figure}

The 15-puzzle is a classic ``sliding tile'' puzzle that was first
designed in 1880 and became very popular in the 1880s~\cite{slocum200615}.
It consists of a $4\times4$
grid containing tiles numbered~$1$ through~$15$ along with one missing tile
(see Figure~\ref{fig:15puzzle}).
The objective of the puzzle is to arrange the tiles so that they are in
ascending order when read from left to right and top to bottom and to
end with the blank tile in the lower right.
The only moves allowed are those that slide a tile adjacent to the blank space
into the blank space.  Half of the possible starting positions are solvable~\cite{archer1999modern}
and the hardest legal starting positions require eighty moves to complete~\cite{brungger1999parallel}.

Our encoding of the 15-puzzle is more complicated because unlike the other problems
we've considered, a solution to the 15-puzzle is not static.  In other words, our encoding must be
able to deal with the state of the puzzle changing over time.  To do this we use the variables
$S_{i,j,n,t}$ to denote that the entry at $(i,j)$ contains tile~$n$ at timestep~$t$.
Here $1\leq i,j\leq 4$, $1\leq n\leq 16$ (we let~$16$ denote the blank tile), and $0\leq t\leq 80$
since each instance requires at most~$80$ moves to complete.

The board does not permit two tiles to occupy the same location at the same time; these constraints
are of the form $S_{i,j,n,t}\limp\lnot S_{i,j,m,t}$ for each tile numbers $n\neq m$, valid indices~$i$
and~$j$, and valid timesteps~$t$.  Next we need to generate constraints that tell the SAT solver how
the state of the board can change from time~$t$ to time $t + 1$.
There are two cases to consider, depending on if a square (or an adjacent square) contains the blank tile.

The easier case is when a square $(i,j)$ and none of the squares adjacent to that square are blank.
In that case, the rules of the puzzle imply that the tile in square $(i,j)$ does not change.
We define the function $\doesNotChange(i,j,t)$ to be the constraint that says that the tile in square 
$(i,j)$ does not change at time~$t$; these constraints are of the form
$\bigwedge_{n=1}^{16}(S_{i,j,n,t}\liff S_{i,j,n,t+1})$.
We also use the function $\adj(i,j)$ to denote the squares adjacent to $(i,j)$ and define
$\notEqualOrAdjacentToBlank(i,j,t)$ to be $\lnot S_{i,j,16,t}\land\bigwedge_{(k,l)\in\adj(i,j)}\lnot S_{k,l,16,t}$.
The static transition constraints are of the form
\[ \notEqualOrAdjacentToBlank(i, j, t) \limp \doesNotChange(i, j, t) \]
for all valid squares $(i, j)$ and timesteps~$t$.

The harder transition case is when a square $(i,j)$ contains the blank tile.
In this case we need to encode the fact that the blank tile will switch positions with
exactly one of the squares adjacent to square $(i,j)$.
If the tile on square $(i, j)$ switches positions with the square $(k,l)$ at time~$t$
this can be encoded as the constraint 
$\bigwedge_{n=1}^{16}(S_{i,j,n,t}\liff S_{k,l,n,t+1})$.
We also need to enforce that all squares adjacent to $(i,j)$ other than $(k,l)$
do not change; these constraints are of the form 
$\bigwedge_{(x,y)}\doesNotChange(x,y,t)$ where $(x,y)$ is adjacent to $(i,j)$ but
not equal to $(k,l)$.
Let $\oneTileMoved(i,j,k,l,t)$ denote the conjunction of the above two constraints.
Then the slide transition constraints are of the form 
\[ S_{i,j,16,t} \limp \bigvee_{(k,l)\in\adj(i,j)}\oneTileMoved(i,j,k,l,t) \]
for all valid squares $(i, j)$ and timesteps~$t$.

The constraint $\boardSolved(t)$ that says the board is solved at timestep~$t$ can be encoded as
$\bigwedge_{i,j=1}^4 S_{i,j,4i+j-4,t}$.
For efficiency reasons we only start looking for solutions with at most 5 moves;
if no solution is found then we look for solutions using at most 10 moves and continue in this
manner until a solution is found.  In other words, we call \texttt{Satisfy} with
the given starting constraints, the constraints of the puzzle as described above,
and the constraint $\bigvee_{t=m-4}^m\boardSolved(t)$ where~$m$ is initialized to~$5$ and then
increased by~$5$ every time no solution is found.

This method was applied to the starting configuration from Figure~\ref{fig:15puzzle}.
It found that no solutions with at most 5 moves exist in 1.5 seconds, no solutions with at most 10
moves exist in 2.8 seconds, and found a solution with 15 moves in 6.3 seconds.  It was also able to
solve puzzles requiring up to 40 moves in 20 minutes.  While this is not competitive with
dedicated solvers for the 15-puzzle, it requires no knowledge beyond
the rules of the game and makes an interesting example of how to
push SAT solvers to their limits.

\section{Conclusion}\label{sec:conclusion}

In this paper we've demonstrated how to solve a variety of problems and puzzles
using the computer algebra system Maple and its SAT solver MapleSAT~\cite{liang2017empirical}.
We discussed a number of encodings and ways for improving those encodings,
e.g., by using symmetry breaking (as in Section~\ref{sec:latin})
or by using auxiliary variables (as in Section~\ref{sec:clique}).
We also took advantage of Maple's ability to solve SAT problems not
encoded in conjunctive normal form
in Sections~\ref{sec:nqueens}, \ref{sec:latin}, and~\ref{sec:15puzzle}.
Maple code for all the examples covered in this paper (including code to read the output
of \texttt{Satisfy} and generate the figures included in this paper) are available for download
from the Maple Application Center~\cite{brightapps}.

The implementations presented in this paper can be considered examples of
\emph{declarative programming} where the programmer focuses on describing
the problem but not the solution---the computer automatically decides the
best way to solve the problem.
This is in contrast to \emph{imperative programming}
where a programmer needs to describe precisely
\emph{how} the computation is to take place.
An advantage of declarative programming is that the programmer does not
need to worry about specifying a potentially complicated search algorithm.
However, a disadvantage of declarative programming
is that it lacks the kind of detailed control over the method of solution that
can be required for optimally efficient solutions.  Furthermore,
not all problems are naturally expressed in a declarative way.  

As we saw in the maximum clique problem, sometimes declarative solutions can outperform
imperative solutions.  This also occurs in the \emph{graph colouring}
(or \emph{chromatic number}) problem of colouring the vertices
of a graph using the fewest number of colours subject to the constraint
that adjacent vertices are coloured differently.
For example, prior to Maple 2018 the \texttt{ChromaticNumber}
function required several hours to find a minimal colouring of the $8\times8$ queens graph
but a SAT encoding can solve this problem in under~10 seconds~\cite{brightapps}.
The SAT approach is available in Maple 2019 using the \texttt{method=sat}
option of \texttt{ChromaticNumber}.

This somewhat unconventional manner of using Maple is not applicable to all
problems but we hope the examples in this paper have convinced the reader that
SAT solvers are more useful and powerful than they might at first appear.
With its extensive logic functionality and convenient method of expressing
logical constraints, Maple is an ideal tool for experimenting with SAT solvers
and logical programming.

\bibliographystyle{splncs04}
\bibliography{maple}

\end{document}